\pdfoutput=1
\documentclass[11pt]{article}

\usepackage[final]{acl}

\usepackage{times}
\usepackage{latexsym}

\usepackage[T1]{fontenc}
\usepackage[utf8]{inputenc}

\usepackage{microtype}

\usepackage{inconsolata}

\usepackage{graphicx}
\usepackage{multicol}
\usepackage{multirow}
\usepackage{amsmath}
\usepackage{amssymb}
\usepackage{subcaption}

\title{Beyond Logit Lens: Contextual Embeddings for Robust Hallucination Detection \& Grounding in VLMs}

\author{
 \textbf{Anirudh Phukan\textsuperscript{3}},
 \textbf{Divyansh\textsuperscript{1*}},
 \textbf{Harshit Kumar Morj\textsuperscript{2*}},
 \textbf{Vaishnavi\textsuperscript{1*}},
\\
 \textbf{Apoorv Saxena\textsuperscript{3}},
 \textbf{Koustava Goswami\textsuperscript{3}},
\\
\\
 \textsuperscript{1}IIT Kanpur,
 \textsuperscript{2}IIT Bombay,
 \textsuperscript{3}Adobe Research
\\
 \small{
   \textbf{Correspondence:} \href{mailto:phukan@adobe.com}{phukan@adobe.com}
 }
}

\begin{document}
\maketitle
\renewcommand{\thefootnote}{} % Remove footnote numbering
\footnotetext{Authors marked with * were interns at Adobe Research when this work was done.}
\begin{abstract}
The rapid development of Large Multimodal Models (LMMs) has significantly advanced multimodal understanding by harnessing the language abilities of Large Language Models (LLMs) and integrating modality-specific encoders. However, LMMs are plagued by hallucinations that limit their reliability and adoption. While traditional methods to detect and mitigate these hallucinations often involve costly training or rely heavily on external models, recent approaches utilizing internal model features present a promising alternative. In this paper, we critically assess the limitations of the state-of-the-art training-free technique, the logit lens, in handling generalized visual hallucinations. We introduce \textit{ContextualLens}, a refined method that leverages contextual token embeddings from middle layers of LMMs. This approach significantly improves hallucination detection and grounding across diverse categories, including actions and OCR, while also excelling in tasks requiring contextual understanding, such as spatial relations and attribute comparison. Our novel grounding technique yields highly precise bounding boxes, facilitating a transition from Zero-Shot Object Segmentation to Grounded Visual Question Answering. Our contributions pave the way for more reliable and interpretable multimodal models. 
\end{abstract}

\section{Introduction}

\begin{figure}[htbp]
    \centering
    \includegraphics[width=0.48\textwidth]{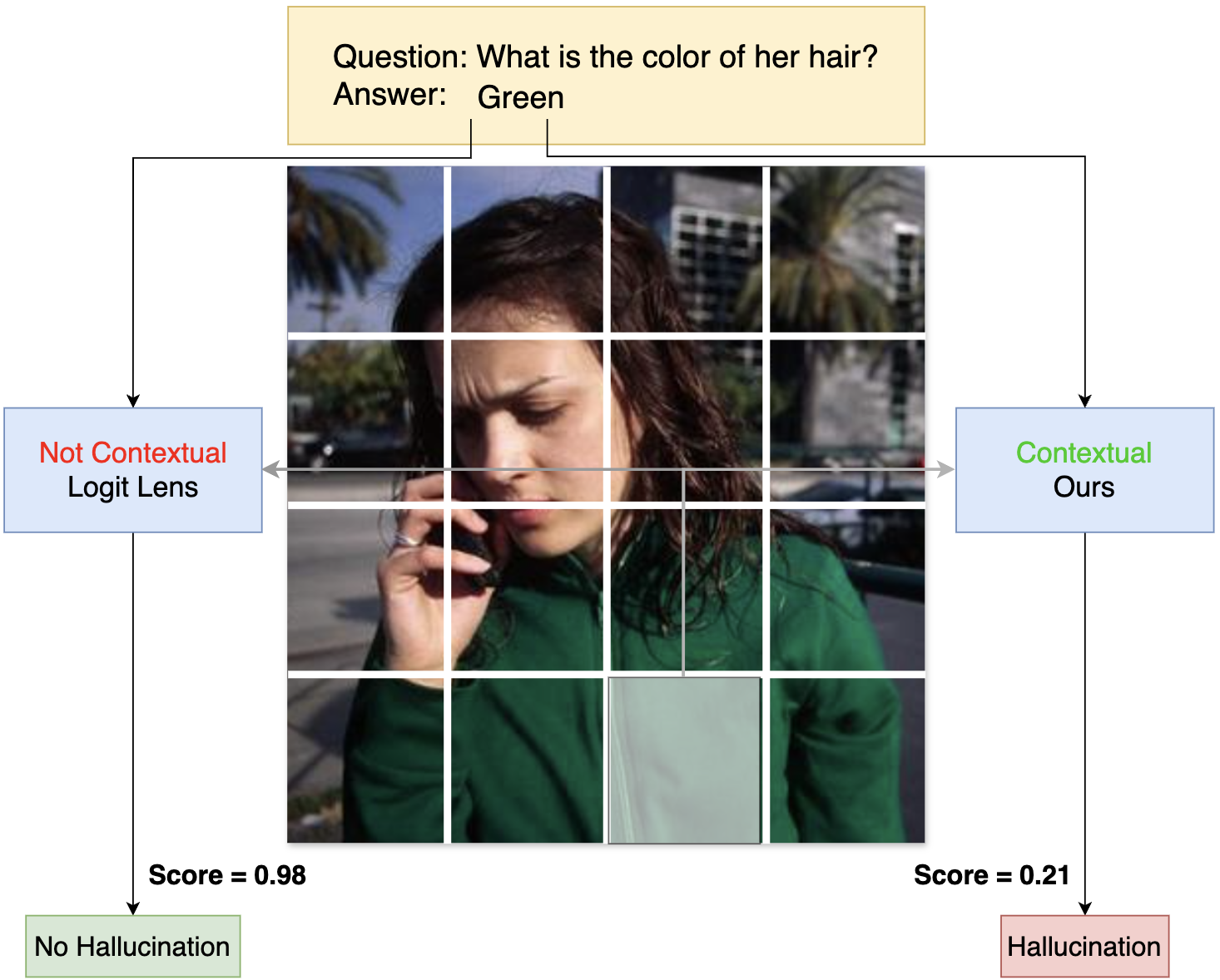}  
    \caption{\textbf{Visual Hallucination Detection}. Logit lens, which only verifies the presence of the token "Green" in the image, mistakenly considers the answer correct. In contrast, \textit{ContextualLens} assigns a low score to "Green" as it correctly contextualizes the color to the jacket rather than the hair.}
    \label{fig:teaser}
\end{figure}

\begin{figure*}[t]
    \centering
    \includegraphics[width=0.98\textwidth]{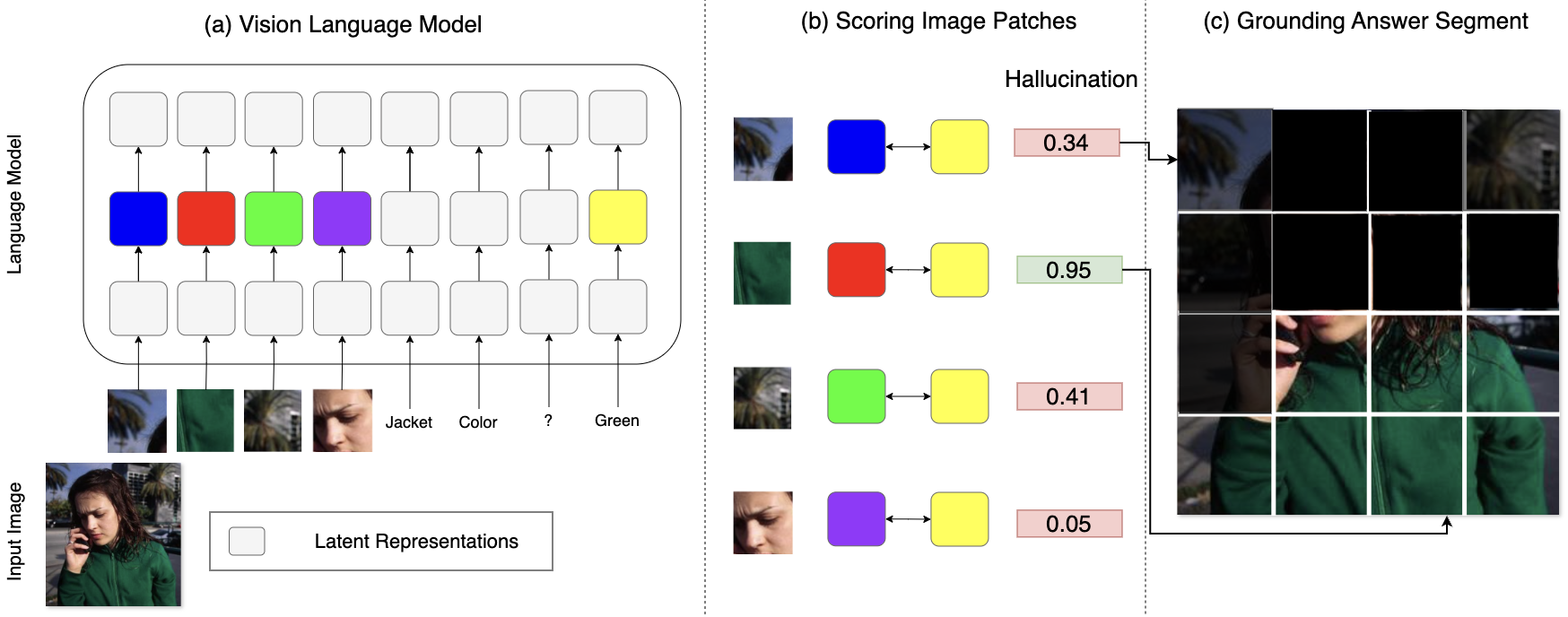}  
    \caption{(a) We extract latent representations of image patches and answer tokens from intermediate layers, (b) compute the cosine similarity between each image patch and the average embedding of the answer tokens to score patches for hallucination detection, and (c) ground the answer in specific image patches.}
    \label{fig:pipeline}
\end{figure*}

Recent advancements in multimodal understanding have been significantly driven by Large Multimodal Models (LMMs), which capitalize on the language capabilities of Large Language Models (LLMs) and integrate modality-specific understanding by training adapters that connect LLMs to pre-trained modality-specific encoders \cite{jin2024efficient}. However, critical issues inherent to LLMs, such as the tendency to produce highly confident incorrect answers—known as hallucinations—are also transferred to LMMs \cite{bai2024hallucination}. Moreover, LMMs introduce additional hallucinations specific to the integrated modalities \cite{liu2024survey}. Addressing these hallucinations, both by detecting their occurrence and mitigating their effects, as well as by providing evidence to support the generated responses, is crucial for fostering user trust and facilitating the widespread adoption of these technologies \cite{bohnet2022attributed}.

Detecting and mitigating hallucinations has been extensively explored in both language \cite{tonmoy2024comprehensive} and vision domains \cite{liu2024survey}. In the context of language, attribution and citation have been proposed as approaches for providing evidence to support model output \cite{gao2023enabling, huo2023retrieving}. In computer vision, similar methods are typically studied under Grounded Visual Question Answering \cite{zhang2024groundhog, khoshsirat2023sentence}. However, many existing techniques necessitate either training from scratch or fine-tuning \cite{jiang2024interpreting}, and frequently rely on external models such as retrievers or object detectors. These requirements impose significant training costs or result in increased latency during inference, posing challenges to their practical deployment in real-world scenarios. 

Recent studies have increasingly leveraged internal model features, such as latent representations, logits, and attention weights, to tackle the objectives of detecting hallucinations \cite{azaria2023internal, varshney2023stitch} and ensuring the grounding of generated outputs in LLMs \cite{phukan2024peering, qi2024model}. Notably, one such study demonstrates the efficacy of a training-free interpretability technique, known as the logit lens, in identifying and mitigating object hallucinations within Visual-Language Models (VLMs) \cite{jiang2024interpreting}.

The logit lens technique, originally introduced in the context of language models, involves directly mapping intermediate activations to the vocabulary space using the unembedding layer, allowing for an interpretable view of token predictions at different layers \cite{nostalgebraist2020logitlens}. By applying this technique to VLMs, the authors probe each image patch to determine the presence of objects.

However, we identify several fundamental limitations with the application of the logit lens in this context, which hinder its effectiveness in addressing more general forms of visual hallucinations as categorized by \citet{yan2024evaluating}. We contend that the reliance on token embeddings from the unembedding layer—which are neither contextual nor easily combinable to form multi-token concepts—results in its failure to handle more generalized hallucination scenarios. Consequently, we propose leveraging token embeddings from the middle layers, which have been shown to be contextual \cite{phukan2024peering} and effective in representing concepts \cite{wendler2024llamas}.

By employing contextual answer token embeddings, we are able to detect hallucinations in categories that were previously near random performance and improve detection accuracy in other categories. Additionally, this method enables a transition from Zero-Shot Object Segmentation using the logit lens technique to performing the more general Grounded Visual Question Answering task. This advancement underscores the potential of our approach in enhancing the reliability and interpretability of multimodal models while operating in the training-free paradigm.

Our contributions are threefold: 
\begin{itemize}
    \item We investigate the robustness of a \textit{SOTA} training-free VLM Object Hallucination Detection and Segmentation method on VQA datasets. We find that the method does not generalize well to actions and OCR, also completely failing on tasks requiring contextual understanding such as attributes, spatial relations, and comparisons.
    \item We extend the method by introducing \textit{ContextualLens}, which replaces the use of the logit lens with contextual embeddings from the middle layers of the VLM, leading to successful hallucination detection in categories that previously performed near random and improving detection accuracy in other categories. 
    \item We propose a novel grounding technique that returns highly precise bounding boxes, enabling a transition from Zero-Shot Object Segmentation to performing the more general Grounded Visual Question Answering task.
\end{itemize}

\section{Related Work}

\subsection{Hallucination Detection and Mitigation in Large Foundation Models}

Recent advancements in hallucination detection within LLMs have been driven by studying output logits \cite{varshney2023stitch}, activations \cite{chen2024inside}, and latent representations \cite{azaria2023internal, su2024unsupervised}. Despite the progress in detection, predominant strategies for hallucination mitigation encompass Retrieval Augmented Generation \cite{gao2022rarr, peng2023check}, Iterative Prompting \cite{ji2023towards}, Supervised Fine-tuning \cite{tian2023fine}, and Alternative Decoding Strategies \cite{shi2023trusting, chuang2023dola}, requiring specific training or external models. 

Additionally, VLMs necessitate distinct approaches to tackle issues inherent to the vision modality. Effective methods span increasing visual resolution \cite{bai2023qwen, li2024monkey}, integrating segmentation and depth maps \cite{jain2024vcoder}, enhancing connection modules \cite{chen2024internvl}, and optimizing the decoding process \cite{huang2024opera}. \citet{jiang2024interpreting} propose an innovative, training-free method, using the logit lens to probe individual image patch embeddings for the detection and mitigation of object hallucinations. 

\subsection{Attribution and Grounded Visual Question Answering}

In the language domain, citing or attributing generated texts to the sources are commonly studied for building user trust. Methods fall into three major categories: attribution using fine-tuned/trained models \cite{gao2023enabling, sun2022recitation}, attribution using external retrievers or auxiliary models \cite{huo2023retrieving, lee2019latent, ramu2024enhancing, sancheti2024post}, and methodologies that harness internal model features directly \cite{phukan2024peering, qi2024model}. \citet{phukan2024peering} leverage contextual embeddings from the intermediate layers of LLMs to pair answer tokens with document tokens by utilizing embedding similarity. Similarly, \citet{qi2024model} employ KL-divergence measures between logits with and without contextual information to pinpoint context-sensitive output tokens.

In the vision-language domain, the task of providing evidence for model outputs has been extensively explored under Grounded Visual Question Answering (GVQA). Prior to the development of LMMs, GVQA relied on custom, end-to-end trained architectures \cite{tan2019lxmert, zhang2021vinvl} or specialized attention mechanisms \cite{urooj2021found, khoshsirat2023sentence}. More recent advancements have given rise to grounding LMMs \cite{peng2023kosmos, rasheed2024glamm}, which generate segmentation masks or bounding boxes for objects and attributes in related tasks such as Grounded Conversation Generation (GCG), Grounded Image Captioning (GIC), and Reference Expression Segmentation (RES). Notably, \textit{GROUNDHOG} \cite{zhang2024groundhog} is a grounding LMM capable of performing GVQA. 

\textit{ContextualLens} is inspired by the training-free object hallucination detection and mitigation method proposed by \citet{jiang2024interpreting}, but with a crucial distinction. We utilize output token embeddings from intermediate layers instead of employing the logit lens probe. This modification enables us to overcome several limitations of the previous work, particularly by enhancing the representation of multi-token concepts and contextual relationships (\S \ref{method}).

\begin{figure*}[t]
    \centering
    \subfloat[Big Ben]{
        \includegraphics[width=0.48\textwidth]{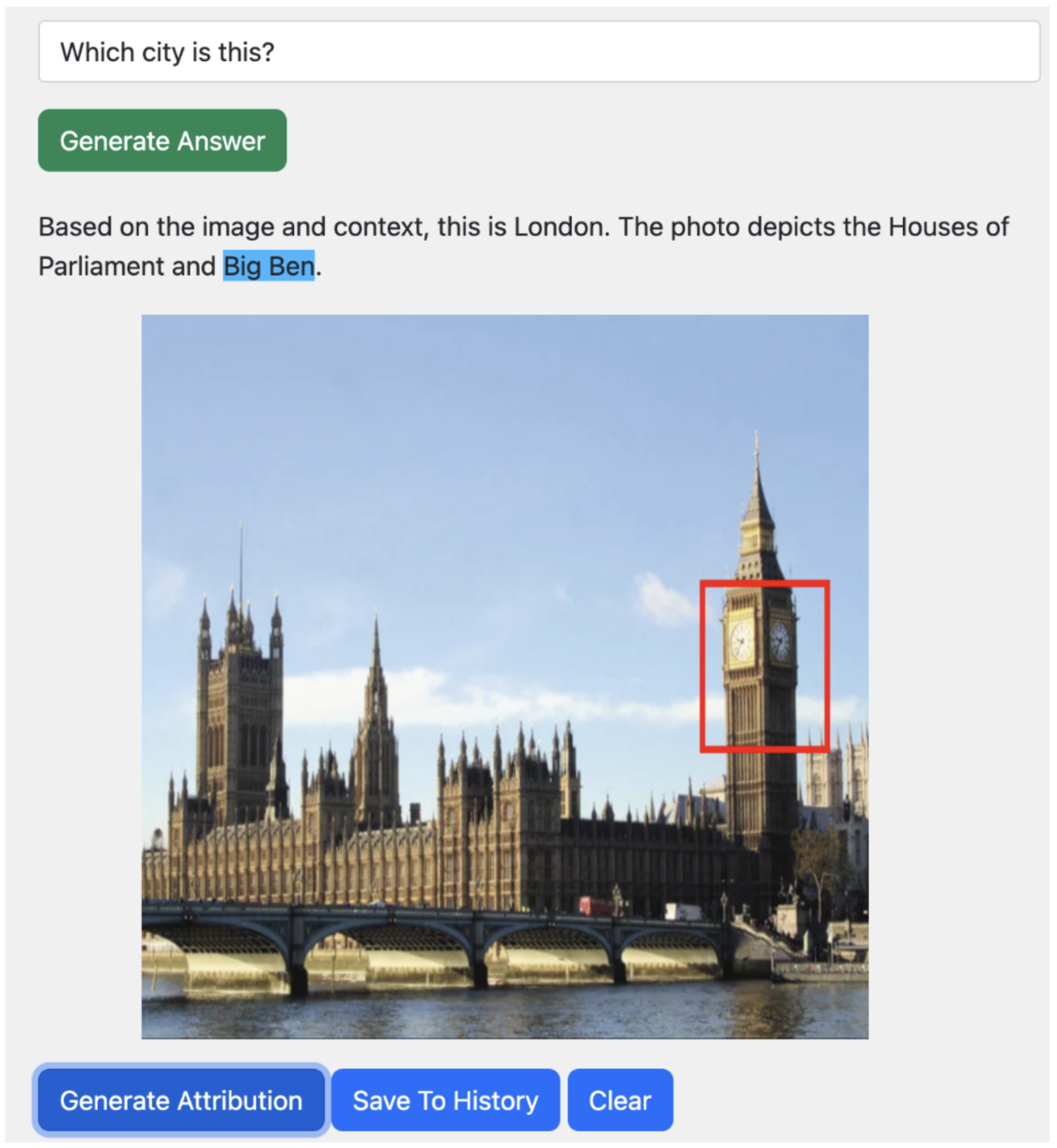}
        \label{fig:bigben}
    }
    \hfill
    \subfloat[Chart]{
        \includegraphics[width=0.46\textwidth]{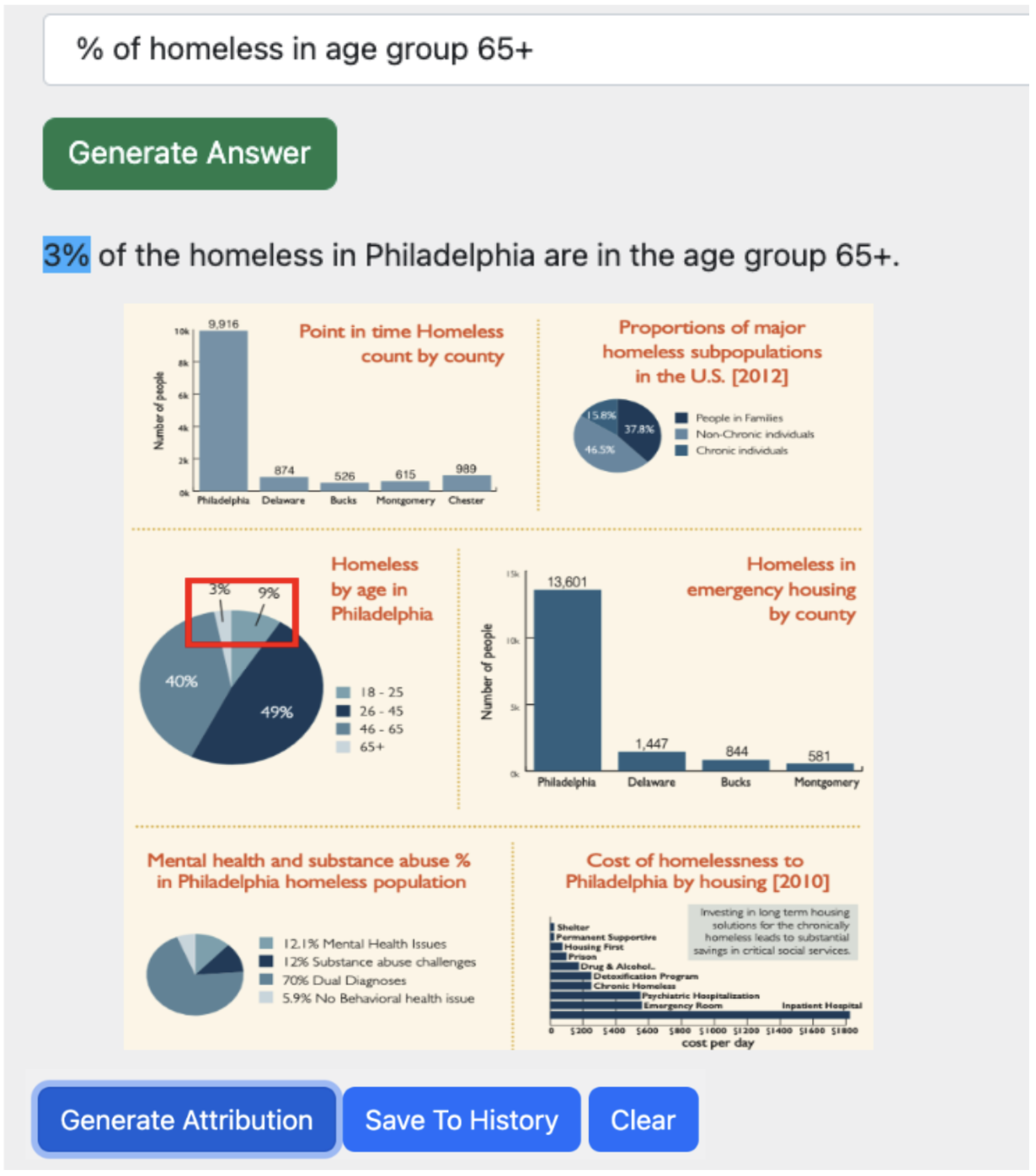}
        \label{fig:chart}
    }
    \caption{Qualitative examples for Grounded Visual Question Answering. Here is a GUI where we pass an image and ask a question. After the answer is generated, we select the span and click "Generate Attribution". Observe, the bounding boxes correctly ground the selected span to the image.}
    \label{fig:qualitative1}
\end{figure*}

\section{Preliminaries}

\subsection{Vision Language Model Architecture}

Vision-Language Models (VLMs) such as Llava \cite{liu2024visual} use the following general recipe to blend visual and textual inputs. Formally, let $\mathcal{M}_{vl}$ denote a VLM, comprising three primary components: a vision encoder $\mathcal{V}$, a connection module $\mathcal{C}$, and a Large Language Model (LLM) $\mathcal{M}_l$. Given an input image $I$, the vision encoder $\mathcal{V}$ processes this image to produce $n$ image features, each corresponding to a distinct patch within the image. These features are then projected by the connection module $\mathcal{C}$ into the $d$-dimensional input embedding space of the LLM $\mathcal{M}_l$. Subsequently, a textual prompt $p$ is concatenated with the projected image features to generate a sequence of $m$ text features. This combined sequence traverses through the $L$ layers of $\mathcal{M}_l$, where each layer refines and integrates the multimodal information. For a given input embedding $x$, we denote the latent representation at layer $l$ as $h_l(x)$. Lastly, at the final layer, these latent representations are mapped back to the vocabulary space using the Unembedding Matrix $W_U \in \mathbb{R}^{|V| \times d}$, where $V$ represents the vocabulary of $\mathcal{M}_l$.

\subsection{Interpreting Latent Representations}
\label{sec:logits}
The logit lens technique involves applying the Unembedding Matrix $W_U$ to intermediate latent representations $h_l(x)$ in order to obtain logit distributions over the entire vocabulary. \citet{jiang2024interpreting} utilized this technique to estimate the probability of individual image patches aligning with specific tokens in the vocabulary by performing $W_U \cdot h_l(t_i)$, where $t_i$ represents the intermediate representation of an image patch.

\citet{phukan2024peering} highlighted how to map answers tokens to document tokens using contextual embeddings from intermediate latent representations. Given a span of tokens, the embedding for the entire span is computed by averaging the embeddings of the individual tokens within that span. 

To compare the similarity between spans, cosine similarity is employed. The scoring enables the effective comparison of contextual embeddings derived from different spans, facilitating the identification of semantically related segments within the text.

\subsection{Grounded Visual Question Answering}

Grounded Visual Question Answering (GVQA) extends traditional VQA by incorporating the explicit requirement for the answers to be justified with visual evidence from the image.  
Given an input image $I$, a textual question $Q$, and a VLM $\mathcal{M}_{vl}$, the model generates an answer $A$ and a grounding map $G$. The grounding map $G$ identifies the regions within the image $I$ that correspond to the elements of the answer $A$. The critical challenge in GVQA lies in ensuring that the grounding map $G$ faithfully represents the visual basis for each part of the generated answer $A$.

\section{Proposed Work}

\subsection{Motivation}

Hallucination detection and mitigation in both language and vision modalities have mostly involved extensive re-training or fine-tuning, substantially increasing computational costs and latency, which hampers their real-world applicability. The logit lens methodology, which probes individual image patches using token embeddings from the unembedding layer, marks a step forward by offering a training-free alternative in the vision-language domain.

However, as demonstrated in our analysis (\S \ref{sec:6.1}), the logit lens method exhibits significant performance degradation when applied to more complex hallucination scenarios, such as those involving attributes, comparisons, and relations among objects. This limitation is primarily due to its reliance on non-contextual embeddings, which are incapable of effectively representing multi-token concepts and contextual elements such as spatial relationships and attribute-based discrepancies.

We aim to address the shortcomings of the logit lens by leveraging middle-layer, contextual embeddings known for their ability to encapsulate richer semantic information. By doing so, we hope to enhance hallucination detection across a more comprehensive range of categories and provide precise visual grounding, while maintaining the computational efficiency and training-free nature of the logit lens.

\subsection{Methodology} \label{method}

\subsubsection{Hallucination Detection}
\label{sec:4.2.1}

To perform hallucination detection, we utilize the contextual embeddings from intermediate layers of the VLM. Our method proceeds as follows:

Given the answer tokens generated by the model, we compute the average embedding at a specific layer $ l_T $, denoted as $ h_{\text{Ans},l_T} $. Formally, let $\{t_1, t_2, \ldots, t_k\}$ be the sequence of answer tokens, and $ h_{l_T}(t_i) $ be the embedding of token $ t_i $ at layer $ l_T $. The average embedding $ h_{\text{Ans}, l_T} $ is computed as:

\[
h_{\text{Ans}, l_T} = \frac{1}{k} \sum_{i=1}^{k} h_{l_T}(t_i)
\]

Next, we evaluate the similarity between the answer embedding $ h_{\text{Ans}, l_T} $ and each image patch embedding from a specific layer $ l_I $. Let $\{p_1, p_2, \ldots, p_n\}$ denote the set of image patches, and $ h_{l_I}(p_j) $ be the embedding of patch $ p_j $ at layer $ l_I $. We compute the score for each patch $ p_j $ as the cosine similarity:

\[
\text{Score}(p_j) = \text{CosineSim}(h_{\text{Ans}, l_T}, h_{l_I}(p_j))
\]

The resulting scores, denoted as $\text{Scores}$, represent the relevance of each image patch with respect to the answer embedding. To determine the confidence for hallucination, we consider the maximum score in $\text{Scores}$:

\[
\text{Confidence}_{\text{max}} = \max(\text{Scores})
\]

A high $\text{Confidence}_{\text{max}}$ indicates a low likelihood of hallucination, as it suggests a strong correspondence between the answer and at least one image patch. Conversely, a low $\text{Confidence}_{\text{max}}$ signals potential hallucination, implying that the answer may not be visually supported by any segment of the image.

\subsubsection{Grounded Visual Question Answering}
\label{sec:4.2.2}

To effectively ground the generated answers in corresponding image regions, we introduce two techniques: a refined version of \cite{jiang2024interpreting} based on contextual embeddings and an alternative approach that directly returns bounding boxes for visual grounding.

\textbf{Basic Technique}: In our first approach, we compute the average embedding of the answer tokens across all layers, denoted as $ h_{\text{Ans}} $, with a shape of $ (L, d) $, where $ L $ represents the number of layers and $ d $ the embedding dimension. Formally, given the sequence of answer tokens $\{t_1, t_2, \ldots, t_k\}$, the answer embedding $ h_{\text{Ans}} $ is computed as follows:

\[
h_{\text{Ans}, l} = \frac{1}{k} \sum_{i=1}^{k} h_{l}(t_i) , \; \scriptstyle l \in \{1, 2, \ldots, L\}
\]

This results in a layer-wise answer embedding $ h_{\text{Ans}} = (h_{\text{Ans}, 1}, h_{\text{Ans}, 2}, \ldots, h_{\text{Ans}, L}) $. Simultaneously, for each image patch $ p_j $, we extract its embedding across all layers, denoted as $ h_{p_j} = (h_{1}(p_j), h_{2}(p_j), \ldots, h_{L}(p_j)) $.

For each layer, we compute the cosine similarity between the layer-specific answer embedding $ h_{\text{Ans}, l} $ and the corresponding image patch embedding $ h_{l}(p_j) $. This gives us a set of scores for each patch across layers: $ \text{Scores}_{l} = \{\text{Score}_{l}(p_1), \text{Score}_{l}(p_2), \ldots, \text{Score}_{l}(p_n)\} $.

The final score for each patch $ p_j $ is determined by taking the maximum score across all layers:

\[
\text{FinalScore}(p_j) = \max_{l \in \{1, \ldots, L\}} (\text{Score}_{l}(p_j))
\]

This method allows us to evaluate the relevance of each image patch to the answer tokens, providing a grounding map based on the similarity score. This map is resized to the original image dimensions.

\begin{table*}[t]
\centering
\small
\begin{tabular}{l|cccc|cccc}
\hline
\multirow{2}{*}{Category} & \multicolumn{4}{c|}{InternlmVL-7B} & \multicolumn{4}{c}{Qwen2VL-7B} \\
\cline{2-9}
 & Random & LL & Out Probs  & CL (Ours) & Random & LL & Out Probs & CL (Ours) \\
\hline
Action & 0.776 & 0.795 & 0.787 & \underline{0.796} & 0.604 & 0.636 & 0.710 & \textbf{0.752} \\
Attribute & 0.796 & 0.786 & 0.820 & \underline{0.825} & 0.812 & 0.830 & 0.839 & \textbf{0.911} \\
Comparison & 0.576 & 0.580 & 0.563 & \textbf{0.623} & 0.548 & 0.558 & 0.567 & \textbf{0.685} \\
Count & 0.856 & 0.898 & \textbf{0.946} & 0.885 & 0.804 & 0.860 & \textbf{0.956} & 0.889 \\
Environment & 0.748 & 0.771 & \underline{0.835} & 0.811 & 0.600 & 0.465 & 0.633 & \textbf{0.682} \\
Relation & 0.656 & 0.668 & \underline{0.755} & \underline{0.755} & 0.592 & 0.572 & 0.647 & \underline{0.655} \\
OCR & 0.740 & 0.769 & \textbf{0.856} & 0.772 & 0.740 & 0.793 & 0.860 & \underline{0.871} \\
\hline
\end{tabular}
\caption{Comparison of mAP scores across different hallucination categories in the HQH dataset. The best-performing method in each category is \textbf{bolded} if it is significantly better than the second-best, and \underline{underlined} if the difference is marginal.}
\label{tab:hallmain}
\end{table*}

\textbf{Bounding Box Technique}: In our second approach, we directly identify bounding boxes that correspond to answer tokens, bypassing the need for thresholding each patch. This approach is more robust for practical applications.

Given the average embedding of the answer tokens at a specific layer $ l_b $, denoted as $ h_{\text{Ans}} $, we aim to identify the most relevant bounding box from a set of potential bounding boxes $ \mathcal{S} $. Let $W$ and $H$ be the number of image patches across the width and height of the image, respectively. Thus, $|\mathcal{S}| = W^2 \times H^2$, representing all possible bounding boxes within the image.

For each bounding box $ s \in \mathcal{S} $, we define the embedding $ h_s $ as the average embedding of all the image patches within the bounding box at layer $ l_b $:

\[
h_s = \frac{1}{|P_s|} \sum_{p_j \in P_s} h_{l_b}(p_j)
\]

where $ P_s $ is the set of patches contained in the bounding box $ s $, and $\mathbf{h}_{l_b}(p_j) $ represents the embedding of patch $ p_j $ at layer $ l_b $.

Next, we compute the cosine similarity between the answer embedding $ h_{\text{Ans}} $ and the embedding of each bounding box $ h_s $. The bounding box $ s^* $ that maximizes this cosine similarity is selected as the grounded region:

\[
s^* = \arg\max_{s \in \mathcal{S}} \text{CosineSim}(h_{\text{Ans}}, h_s)
\]

This method harnesses the average embeddings of the answer tokens and bounding boxes to directly locate the most relevant region within the image that corresponds to the generated answer. 

\section{Experimental Setup}

\subsection{Datasets}
We conducted experiments on three datasets to evaluate methods for both tasks outlined in \S \ref{method}. More details can be found in Appendix \ref{appendix:datasets}.

\noindent{\textbf{Hallucination Detection}}: We experimented on High-Quality Hallucination Benchmark (HQH) \cite{yan2024evaluating} dataset, comprising 4,000 image-question pairs accompanied by ground-truth answers and categorized into eight distinct types of potential hallucination scenarios, allowing us to test for the generalization of methods.

\noindent{\textbf{Grounded VQA}}: Experiments on two datasets TextVQA-X \cite{rao2021first} and VizWiz-G \cite{chen2022grounding} highlight the capability of our method's grounding performance on GVQA task. The datasets have 3,620 and 1,131 image-question pairs respectively. 

\subsection{Metrics}

We use the Mean Average Precision (mAP) metric for hallucination detection and plot the Precision-Recall Curve to evaluate the grounding aspect of GVQA. More details can be found in Appendix \ref{appendix:metrics}.

\subsection{Baselines}

\subsubsection{Hallucination Detection}
\label{sec:5.3.1}
We employ two primary baselines for comparative evaluation. The first baseline leverages the maximum probability across the generated answer tokens, referred to as Output Probabilities (Out Probs).  The second baseline is the work by \citet{jiang2024interpreting} (LL). They evaluate the hallucination likelihood by taking the maximum value of the softmax-normalized output token logits over all layers. Logit computation is described in \S \ref{sec:logits}.

\subsubsection{Grounded Visual Question Answering}
\label{sec:5.3.2}
We adapt the method proposed by \cite{jiang2024interpreting} for zero-shot segmentation, which to the best of our knowledge is the only training-free grounding technique for VLMs. We take the mean internal confidence for tokens comprising the answer, similar to how they perform hallucination detection. We resize the set of internal confidence values per image patch back to the size of the image. 

\begin{figure*}[t]
    \centering
    \subfloat{
        \includegraphics[width=0.48\textwidth]{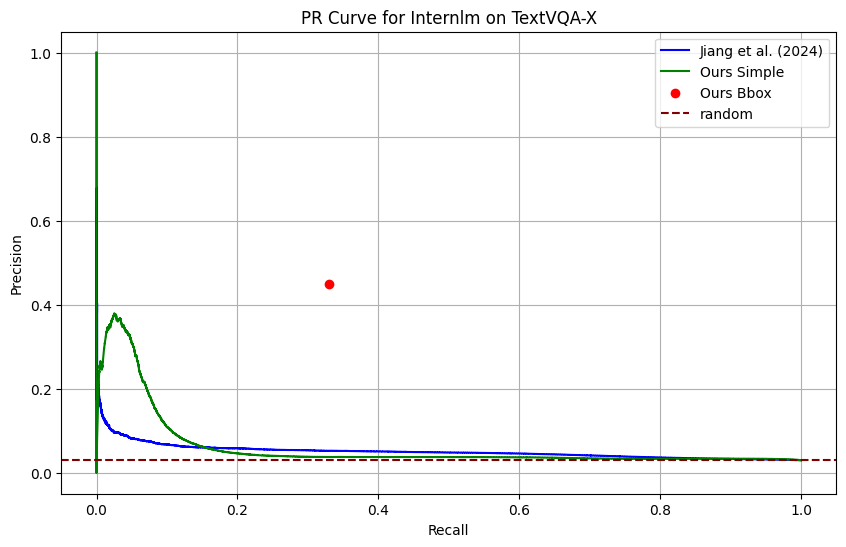}
        \label{fig:internlm_textvqa}
    }
    \hfill
    \subfloat{
        \includegraphics[width=0.48\textwidth]{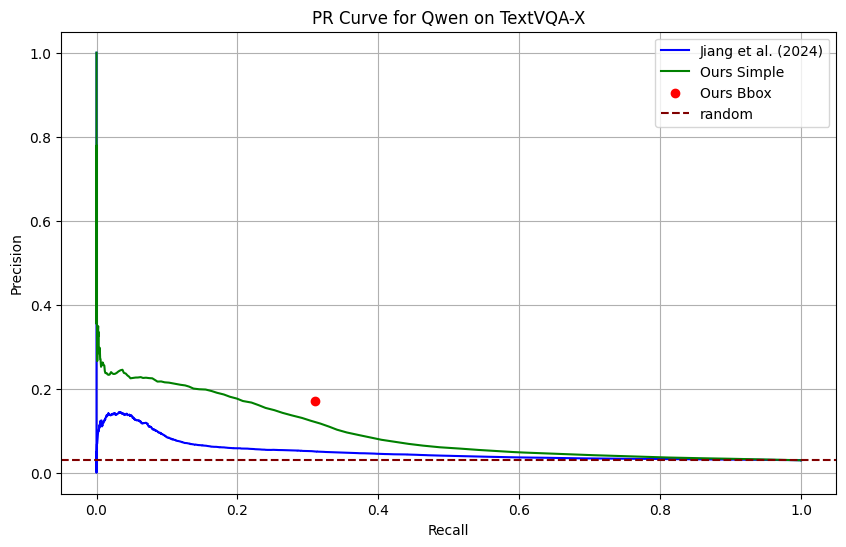}
        \label{fig:qwen_textvqa}
    }
    \caption{PR Curves on the TextVQA-X dataset.}
    \label{fig:prtextvqa}
\end{figure*}

\begin{figure*}[t]
    \centering
    \subfloat{
        \includegraphics[width=0.48\textwidth]{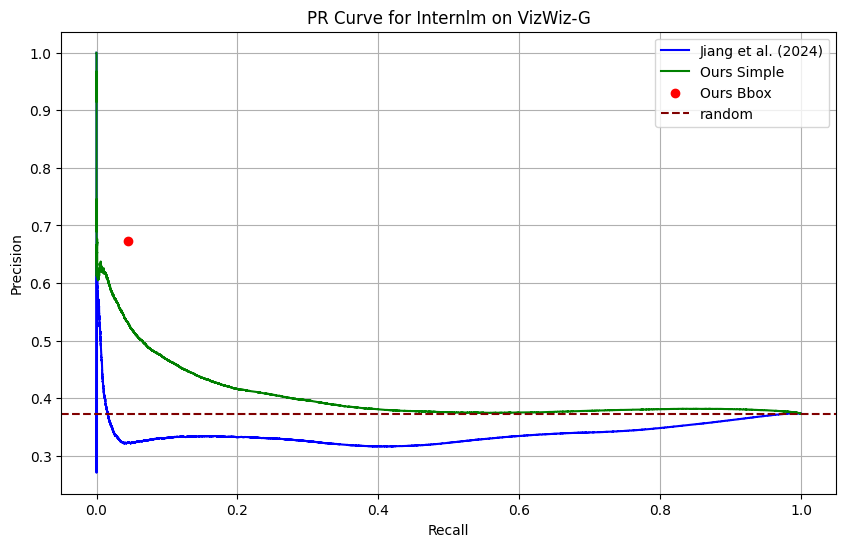}
        \label{fig:internlm_vizwiz}
    }
    \hfill
    \subfloat{
        \includegraphics[width=0.48\textwidth]{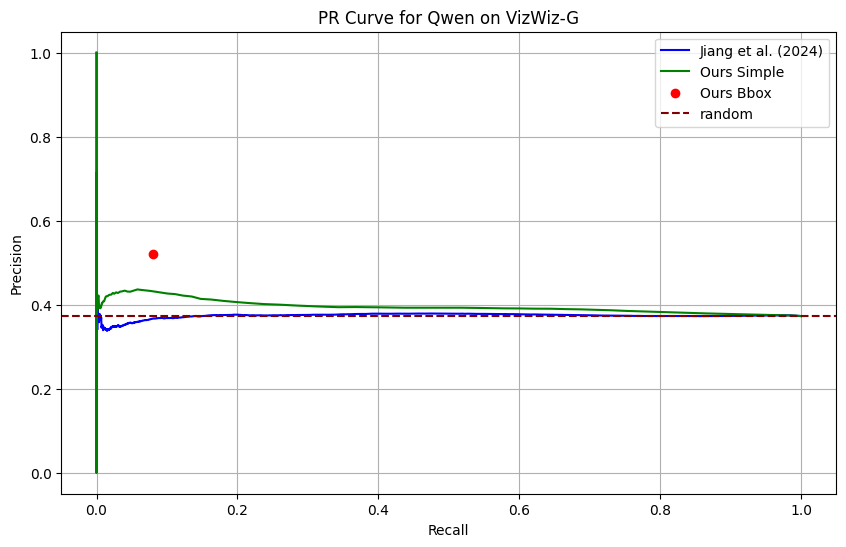}
        \label{fig:qwen_vizwiz}
    }
    \caption{PR Curves on the VizWiz-G dataset.}
    \label{fig:prvizwiz}
\end{figure*}

\section{Results and Analysis}

We perform experiments for our approach with the following VLMs: \texttt{Qwen2-VL-7B} \cite{wang2024qwen2} and \texttt{internlm-xcomposer2-vl-7b} \cite{dong2024internlm}. All of our experiments were run on a A100 machine with 4 80GB GPUs.

\subsection{Hallucination Detection}
\label{sec:6.1}
The performance of \textit{ContextualLens} (\S \ref{sec:4.2.1}) and the baseline methods (\S \ref{sec:5.3.1}) on the HQH dataset is summarized in Table \ref{tab:hallmain}. Each category comprises 500 image-question pairs. To select the best layer combination for our method, we created a test and validation split with 250 examples each. The scores in Table \ref{tab:hallmain} are reported for the test split. Further details on how to perform layer selection in real-world settings, including a robustness analysis using adversarial validation, are provided in Appendix \ref{appendix:layer_sel}.

Notably, the logit lens technique, although effective for object hallucination detection as shown by \cite{jiang2024interpreting}, performs nearly at random in the Attribute, Comparison, and Relation categories across both models. Further investigation into the types of questions in these categories reveals that the Attribute category often involves questions about the color of objects, the Comparison category includes questions comparing two objects (e.g., which is larger), and the Relation category consists of questions about objects' spatial relations. The logit lens technique falls short in these scenarios because it relies on non-contextual embeddings and thus can only determine the presence of objects or features, not their spatial relationships or comparative attributes.

For instance, in the Attribute category, if the question is about the color of an object (e.g., "What is the color of the ball?") and the model answers "blue," the logit lens can only verify the presence of the color blue in the image but not whether the color pertains to the ball or another object like the sky. This underscores the limitation of using non-contextual embeddings for complex hallucination detection tasks.

Our method, which employs contextual embeddings, fares significantly better in these categories compared to the logit lens approach and often outperforms even output probabilities, sometimes by a substantial margin. We refer readers to Figures \ref{fig:attribute} and \ref{fig:comparison} for some qualitative examples.

For categories such as Action and OCR, where grounding actions (e.g., "Looking at the Television") or text (e.g., "Melrose Ave.") is necessary, the logit lens technique performs better than random but still falls short compared to output probabilities. This is likely due to the logit lens's poor handling of multi-token objects. While objects which are usually single token may align relatively well, the complexity of representing actions or texts across multiple tokens poses challenges. Our approach, leveraging middle-layer embeddings known to better represent multi-token concepts \cite{wendler2024llamas}, consistently outperforms the logit lens and, in many instances, surpasses output probabilities. See Figure \ref{fig:ocr} for a qualitative example.

Interestingly, output probabilities excel in the Count category compared to both \textit{ContextualLens} and the logit lens. This category involves questions about the number of instances of a specific object. The inherent challenge for our approach and the logit lens is that visual elements often have low semantic overlap with numerical tokens (e.g., a "car" in the image is unlikely to semantically match "two" in the text). Consequently, output probabilities provide more accurate hallucination detection in such scenarios. A failure case is illustrated in Figure \ref{fig:count}.

\subsection{Grounded Visual Question Answering}
The performance of our methods (\S \ref{sec:4.2.2}) and the baseline method (\S \ref{sec:5.3.2}) on the TextVQA-X dataset and VizWiz-G dataset is summarized in Figure \ref{fig:prtextvqa} and \ref{fig:prvizwiz} respectively.

From the plots, it is evident that our adaptation of \citet{jiang2024interpreting}'s method for object segmentation, by replacing the logit lens with contextual middle layer representations, consistently leads to better grounding for VQA across the tested models and datasets. The TextVQA-X dataset is analogous to the OCR category in the HQH dataset, while VizWiz-G encompasses attributes, OCR, and general object segmentation. Our method's improved grounding performance highlights the utility of using contextual and conceptual embeddings, which also elucidates why we achieve superior hallucination detection.

An essential use case of grounded VQA is not only to understand what image features contribute to answer generation but also to guide users to the general location of the evidence within the image. This need parallels the concept of attribution explored in LLMs and textual domains. We term this broader application in VLMs as multimodal attribution, where precision is more critical than recall. Users prioritize accurate identification of the general evidence location over highlighting every relevant pixel. This is precisely where our bounding box method proves valuable.

As illustrated in Figures \ref{fig:prtextvqa} and \ref{fig:prvizwiz}, the precision of the bounding box method is often significantly higher than what can be achieved by navigating the PR-Curve for the basic version. This precision makes it particularly useful for multimodal attribution, providing users with accurate general locations of evidence, which they can then verify. The higher precision and comparatively lower recall is because we identify the optimal bounding box, which perfectly encloses the relevant evidence. The recall is higher on TextVQA-X since we ground text, and the optimal bounding box often encloses the entire textual element. In VizWiz-G, however, questions about object attributes like color could be satisfied by highlighting a small patch of that color on the object, leading to a lower recall.

\subsubsection{Qualitative Examples}

Some qualitative results of our bounding box visual grounding technique are showcased in Figure \ref{fig:qualitative1}. For instance, in Figure \ref{fig:bigben}, we ground the phrase "Big Ben" to its corresponding location in the image. Our method effectively highlights the relevant portion by leveraging the latent conceptual representation of "Big Ben" intrinsic to the LMM, facilitating the mapping between textual descriptions and visual concepts. Another example is depicted in Figure \ref{fig:chart}. Remarkably, our method proficiently grounds answers within charts and infographics, a novel capability within the community to the best of our knowledge, achievable in a training-free manner. Additional qualitative examples are available in Appendix \ref{sec:add_qual}.

\section{Discussion \& Conclusion}

In this paper, we presented \textit{ContextualLens}, an approach for detecting hallucinations in LMMs by leveraging contextual token embeddings from intermediate layers. We identified limitations in the state-of-the-art training-free technique, the logit lens, notably its poor handling of complex visual hallucinations involving attributes, comparisons, and spatial relations due to its reliance on non-contextual embeddings. Our proposed method employs contextual embeddings known to capture richer semantic information, improving hallucination detection across diverse categories.

Our experimental results on the HQH benchmark showed that \textit{ContextualLens} enables detection in categories that previously performed near random and improves detection for multi-token concept representation. Furthermore, by introducing a novel grounding technique with highly precise bounding boxes, we advanced Zero-Shot Object Segmentation to a more general GVQA task, validating our approach's effectiveness on TextVQA-X and VizWiz-G datasets.

Our method's precision and training-free paradigm circumvent the computational costs of re-training or fine-tuning, promoting its practical applicability in real-world multimodal attribution tasks. Our contributions pave the way for more reliable and user-trustworthy LMMs.

\section{Limitations \& Future Work}

While our work significantly advances the state-of-the-art in training-free hallucination detection and grounding in LMMs, it has some limitations that provide exciting avenues for future research. 

Firstly, the validation of both hallucination detection and grounding has been conducted primarily on factual short answer VQA datasets. Extending this to more diverse and complex datasets remains an area for exploration. Secondly, \textit{ContextualLens} is outperformed by output probabilities in the Count category. Future work could explore extensions to the method for abstractive scenarios such as counting. Lastly, although our method excels at highlighting precise evidence regions sufficient for human verification, improving recall remains an open challenge. Subsequent research could investigate leveraging our scoring mechanisms as priors for pre-trained segmentation models or improving the evaluation metrics for grounded visual question answering.

\bibliography{main}
\clearpage

\newpage
\appendix

\section{Datasets}
\label{appendix:datasets}
\noindent\textbf{High-Quality Hallucination Benchmark} (HQH) \cite{yan2024evaluating}: This dataset comprises 4,000 image-question pairs accompanied by ground-truth answers and is categorized into eight distinct types of potential hallucination scenarios: Attribute, Action, Counting, Environment, Comparison, Relation, OCR, and Existence. The questions in HQH are open-ended, eliciting concise and direct answers. Evaluation within this benchmark is conducted using GPT-3.5 prompting, which has demonstrated high reliability and validity. This benchmark allows us to assess the generalization capabilities of our proposed strategies across multiple hallucination categories. Notably, we exclude the existence category from our evaluation, as our focus is on determining the presence of objects in the image, rather than assessing exhaustive object enumeration.

\noindent\textbf{TextVQA-X} \cite{rao2021first}: This dataset features human-annotated multimodal explanations, including ground truth segmentation maps and multiple references for textual explanations containing text within images. For our experiments, we utilize the validation split, which consists of 3,620 image-question pairs related to scene-text, accompanied by corresponding ground-truth answers. This dataset enables the assessment of grounding performance on GVQA task.

\noindent\textbf{VizWiz-G} \cite{chen2022grounding}: This dataset focuses on visually grounding answers to visual questions posed by individuals with visual impairments. Another GVQA dataset, VizWiz-G's validation split includes 1,131 image-question pairs, each paired with a ground-truth answer and a corresponding segmentation mask.

\section{Metrics}
\label{appendix:metrics}

\subsection{Hallucination Detection}

\textbf{Mean Average Precision} (mAP): We measure hallucination detection by framing it as a binary classification task. mAP provides a comprehensive evaluation by considering the precision-recall trade-offs across various threshold values.

\subsection{Grounded Visual Question Answering}

\textbf{Precision-Recall Curve}: To evaluate the grounding aspect of GVQA, we use Precision-Recall (PR) curves. For methods yielding a confidence score for grounding, we compute precision and recall at various threshold levels and plot the corresponding PR curve. This provides insights into the trade-offs between precision and recall across different settings, offering a detailed evaluation of grounding performance. For methods that directly output a segmentation mask, precision and recall are computed, and the corresponding point is plotted on the PR curve for comparison.

\section{Optimal Layer Selection}
\label{appendix:layer_sel}
We use the hallucination detection task to discuss optimal layer selection, as its diverse categories make it ideal for evaluating robustness. The results in Table \ref{tab:hallmain} were obtained using category-specific validation sets (250 examples per category). To test robustness, we conducted experiments with adversarial validation sets, where the category being tested was excluded from the validation data (1,500 examples: 250 × 6 categories). For InternLM-VL, we observed that Image layer embedding 13 and Text layer embedding 27 consistently ranked among the top two combinations across all categories. These results, summarized in Table \ref{tab:layer_sel}, demonstrate that for all categories—except OCR—the performance closely matches that achieved with task-specific validation, indicating the robustness of layer selection. In the absence of task-specific data, a robust method for selecting optimal layers would be to identify those that rank highest during adversarial validation for available categories.

\begin{table}[ht]
    \centering
    % \small
    \begin{tabular}{l|c|c}
        \hline
        \textbf{Category} & \textbf{Task Specific} & \textbf{Adversarial} \\
        \hline
        Action & 0.796 & 0.792 \\
        Attribute & 0.825 & 0.833 \\
        Comparison & 0.623 & 0.638 \\
        Count & 0.885 & 0.896 \\
        Environment & 0.811 & 0.813 \\
        Relation & 0.755 & 0.752 \\
        OCR & 0.772 & 0.744 \\
        \hline
    \end{tabular}
    \caption{InternLM-VL mAP scores on HQH dataset for task-specific and adversarial validation.}
    \label{tab:layer_sel}
\end{table}

\section{Additional Qualitative examples for Grounded Visual Question Answering}
\label{sec:add_qual}

\begin{figure*}[h!]
    \centering
    \subfloat[Question: What is the color of the \textbf{doll}? Answer: Brown]{
        \includegraphics[width=0.48\textwidth]{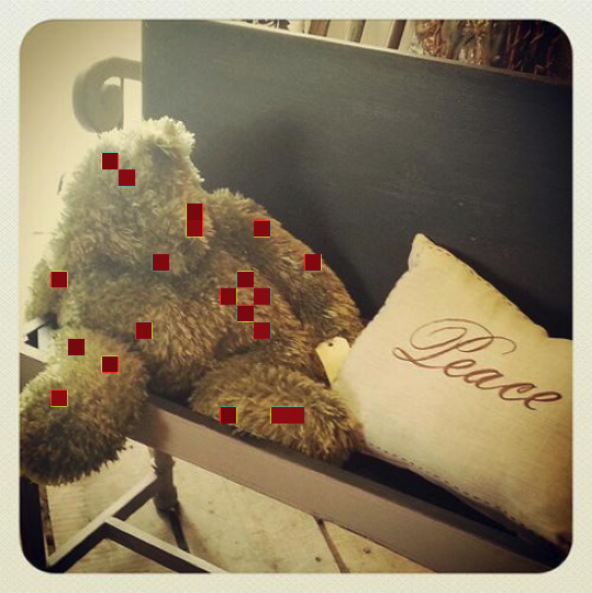}
        \label{fig:attribute_pos}
    }
    \hfill
    \subfloat[Question: What is the color of the \textbf{pillow}? Answer: Brown]{
        \includegraphics[width=0.48\textwidth]{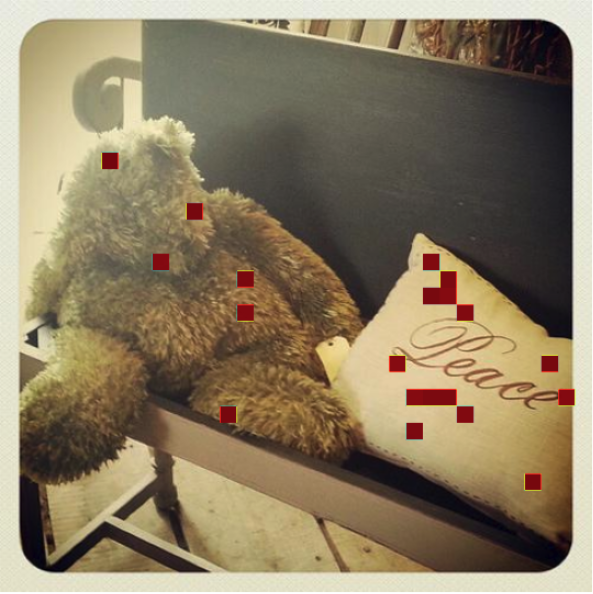}
        \label{fig:attribute_neg}
    }
    \caption{Top 20 image patches while detecting hallucination in \textit{attribute} category using \textit{ContextualLens}. We see that text tokens "Brown" are contextualized by the objects they refer to.}
    \label{fig:attribute}
\end{figure*}

\begin{figure*}[h!]
    \centering
    \subfloat[Question: What is the \textbf{older} boy holding? Answer: Ball]{
        \includegraphics[width=0.48\textwidth]{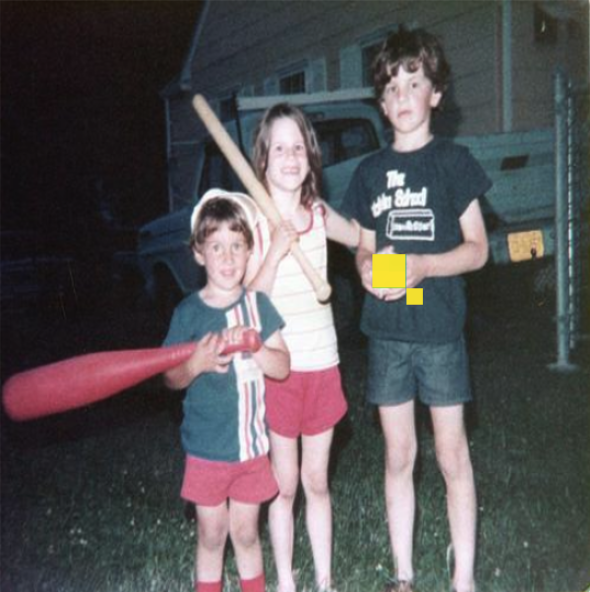}
        \label{fig:comparison_pos}
    }
    \hfill
    \subfloat[Question: What is the \textbf{younger} boy holding? Answer: Ball]{
        \includegraphics[width=0.48\textwidth]{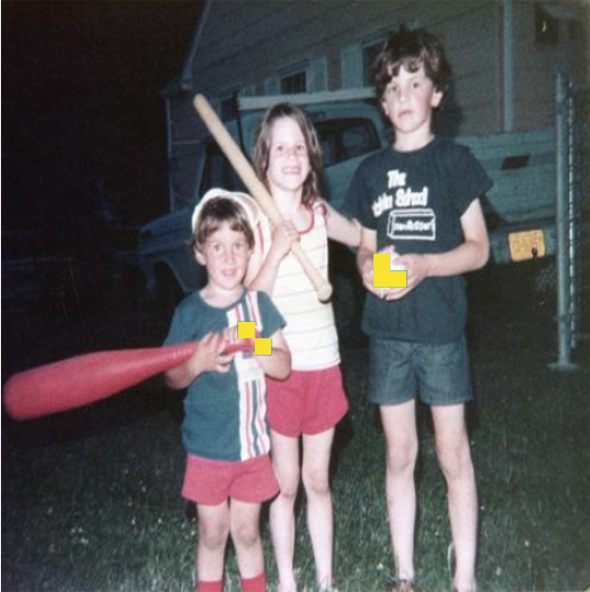}
        \label{fig:comparison_neg}
    }
    \caption{Top 5 image patches while detecting hallucination in \textit{comparison} category using \textit{ContextualLens}. We see that image patches corresponding to the younger boy's hand are also highlighted in the second case.}
    \label{fig:comparison}
\end{figure*}

\begin{figure*}[h!]
    \centering
    \subfloat[Question: What is written on the bus? Answer: \textbf{Michigan}]{
        \includegraphics[width=0.48\textwidth]{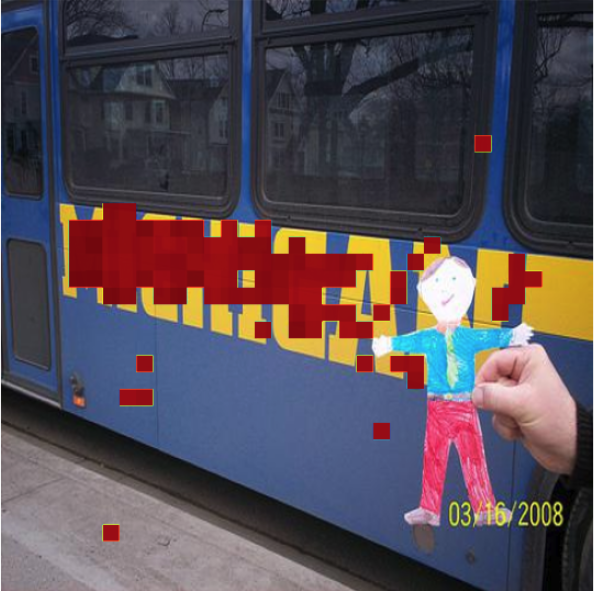}
        \label{fig:ocr_pos}
    }
    \hfill
    \subfloat[Question: What is written on the bus? Answer: \textbf{New York}]{
        \includegraphics[width=0.48\textwidth]{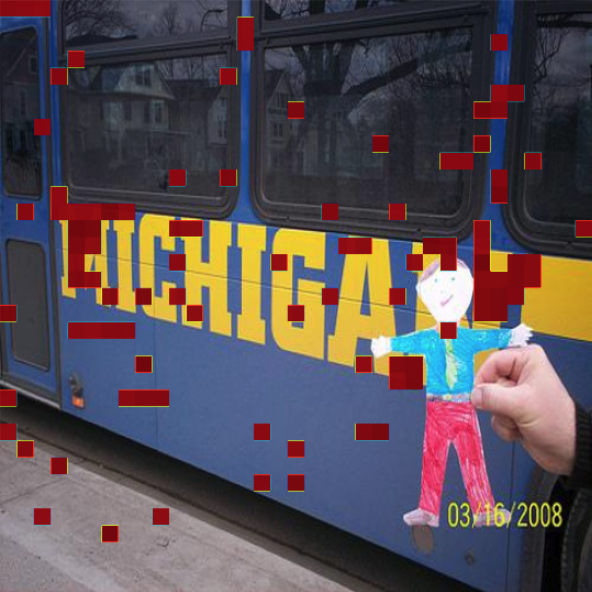}
        \label{fig:ocr_neg}
    }
    \caption{Top 100 image patches while detecting hallucination in \textit{OCR} category using \textit{ContextualLens}. We see that image patches corresponding to multi-token "Michigan" are highlighted in the first case.}
    \label{fig:ocr}
\end{figure*}

\begin{figure*}[h!]
    \centering
    \subfloat[Question: How many people are in this picture? Answer: \textbf{3}]{
        \includegraphics[width=0.48\textwidth]{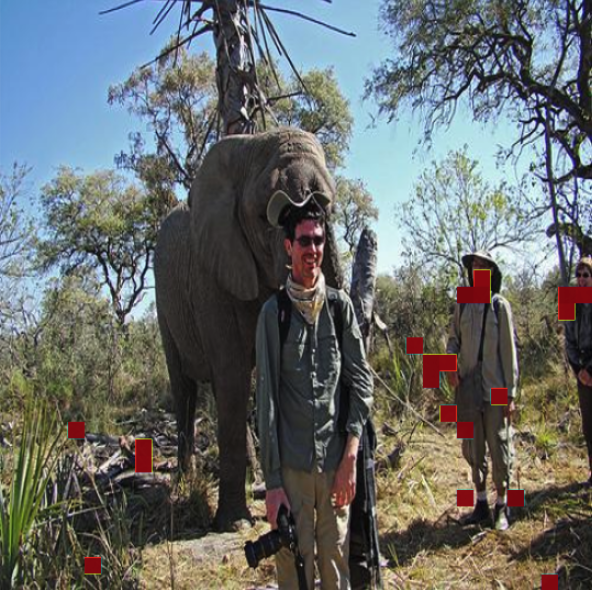}
        \label{fig:count_pos}
    }
    \hfill
    \subfloat[Question: How many people are in this picture? Answer: \textbf{2}]{
        \includegraphics[width=0.48\textwidth]{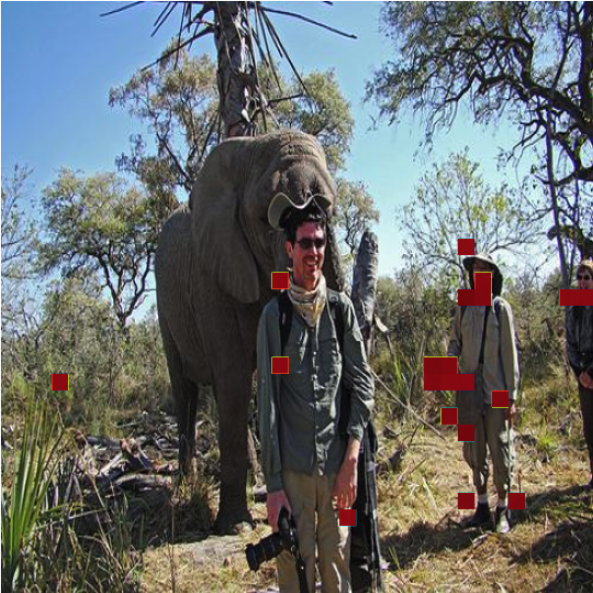}
        \label{fig:count_neg}
    }
    \caption{Top 20 image patches while detecting hallucination in \textit{count} category using \textit{ContextualLens}. We see that image patches highlighted in both cases are similar and the score for the second case is higher (0.891 < 0.892).}
    \label{fig:count}
\end{figure*}

\begin{figure*}[h!]
    \centering
    \subfloat[The image shows a direction board with Chinese text. The red box highlights the Chinese phrase, linking it to the English translation.]{
        \includegraphics[width=0.48\textwidth]{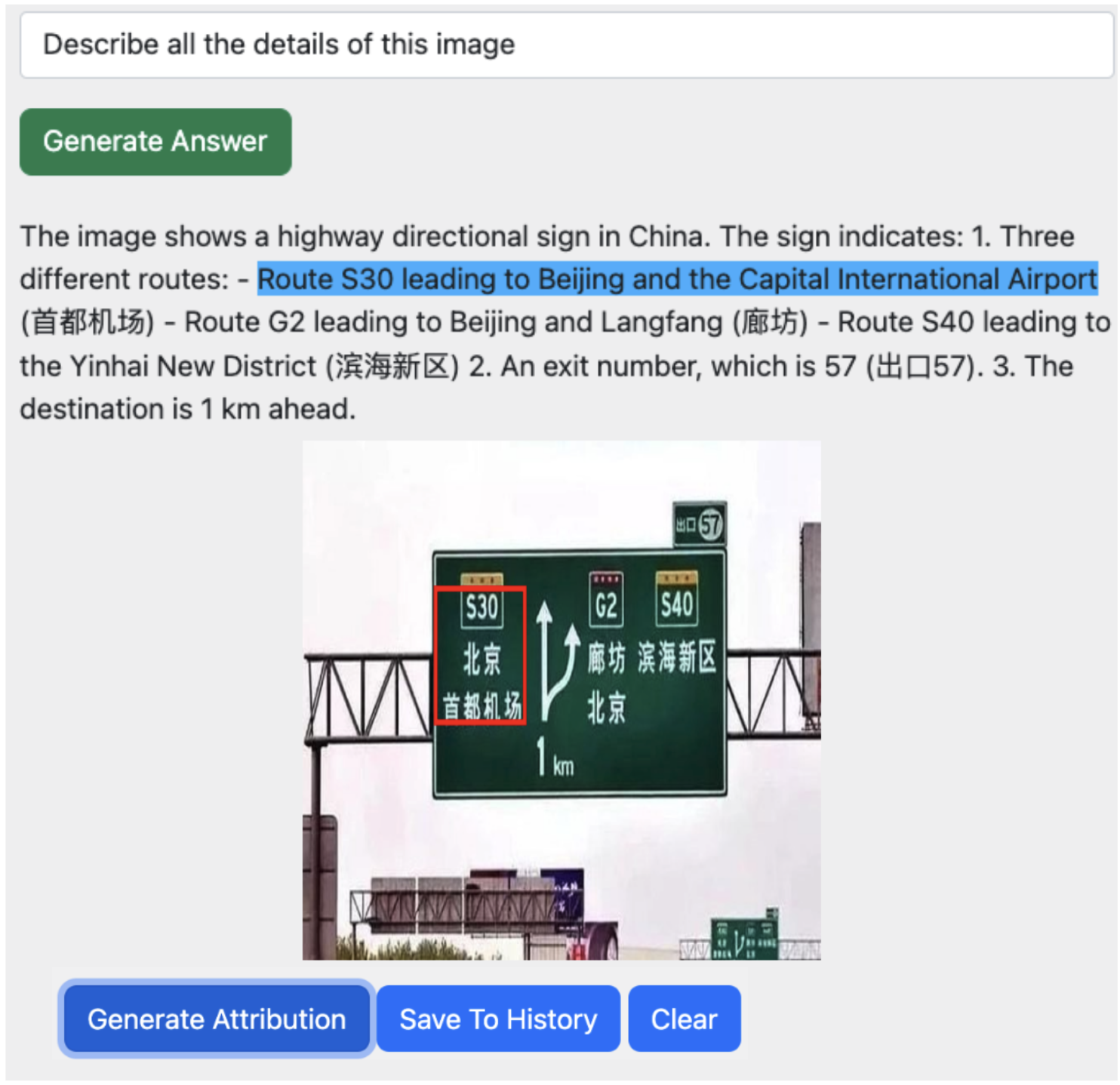}
        \label{fig:route}
    }
    \hfill
    \subfloat[The image shows a hand-filled bank cheque. The red box highlights the word “PAY,” linking it to the phrase “made out to,” which means “to pay someone.”]{
        \includegraphics[width=0.48\textwidth]{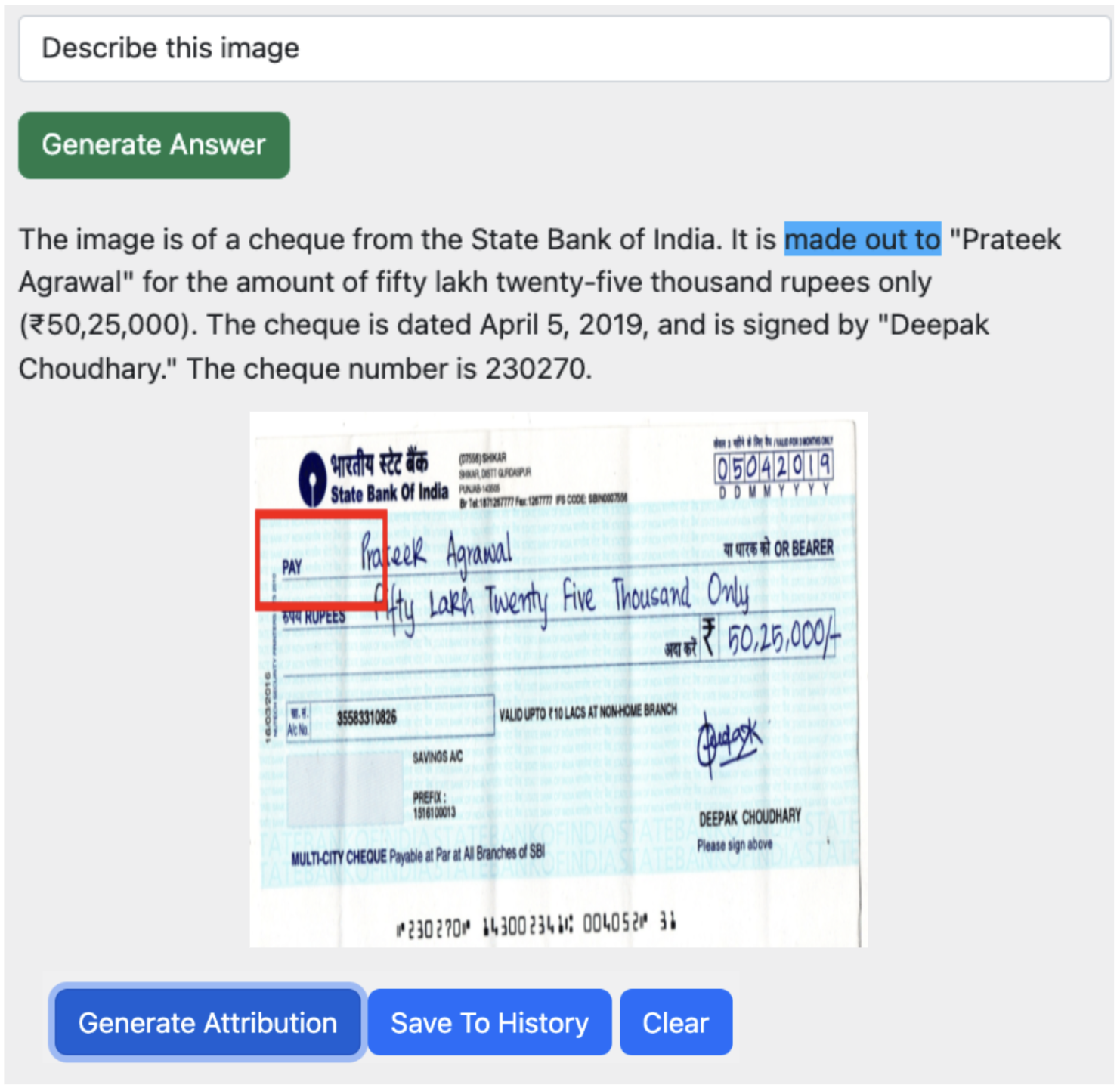}
        \label{fig:cheque}
    }
    \hfill
    \subfloat[The image shows a timetable sheet. The red box highlights the square with the schedule for the date corresponding to the school holiday.]{
        \includegraphics[width=0.48\textwidth]{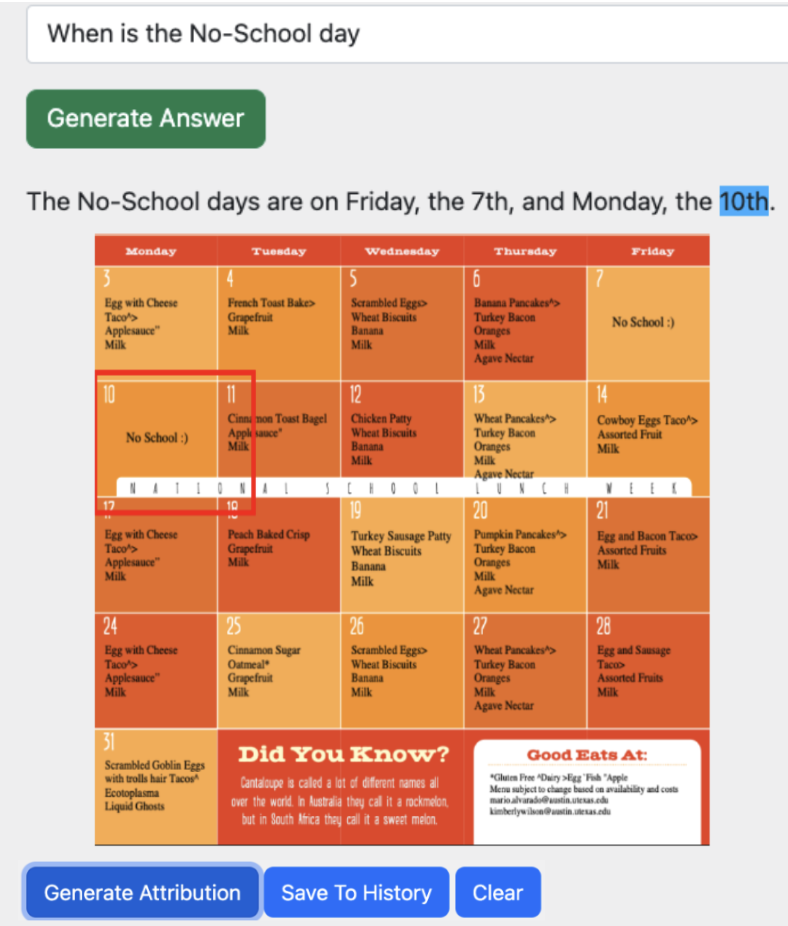}
        \label{fig:timetable}
    }
    \hfill
    \subfloat[The image shows a snapshot of luggage being packed. The red box highlights the notes associated with the packed luggage.]{
        \includegraphics[width=0.47\textwidth]{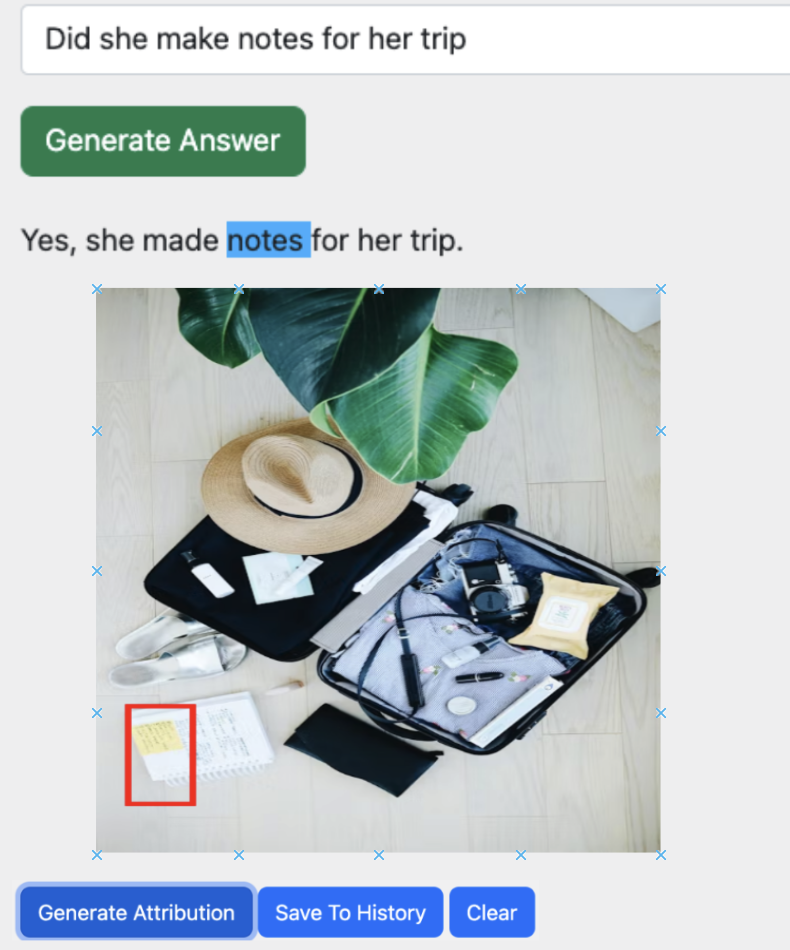}
        \label{fig:notes}
    }
    \caption{Additional qualitative examples for Grounded Visual Question Answering.}
    \label{fig:add_qual}
\end{figure*}

Figure \ref{fig:add_qual} shows some scenarios in which our bounding box based multimodal attribution system could be used.

\end{document}